Oral    Topic: <*Cognitive Approaches for Robotics*>

# From hand to brain and back: Grip forces deliver insight into the functional plasticity of somatosensory processes

**Birgitta Dresp-Langley**

ICube Lab, UMR 7357, Centre National de la Recherche Scientifique CNRS, 67000 Strasbourg, France
E-mail: birgitta.dresp@unistra.fr;

**Summary:** The human somatosensory cortex is intimately linked to other central brain functions such as vision, audition, mechanoreception, and motor planning and control. These links are established through brain learning, and display a considerable functional plasticity. This latter fulfills an important adaptive role and ensures, for example, that humans are able to reliably manipulate and control objects in the physical world under constantly changing conditions in their immediate sensory environment. Variations in human grip force are a direct reflection of this specific kind of functional plasticity. Data from preliminary experiments where wearable wireless sensor technology (sensor gloves) was exploited to measure human grip force variations under varying sensory input conditions (eyes open or shut, soft music or hard music during gripping) are discussed here to show the extent to which grip force sensing permits quantifying somatosensory brain interactions and their functional plasticity. Experiments to take this preliminary work further are suggested. Implications for robotics, in particular the development of end-effector robots for upper limb movement planning and control, are brought forward.

## 1. Introduction

Neuronal activity and the development of functionally specific neural networks in the brain are modulated by sensory signals. The somatosensory cortical network [1] in the primate brain refers to a neocortical area that responds primarily to tactile stimulation of skin or hair. This cortical area is conceptualized in current state of the art [2,3] as containing a single map of the receptor periphery, connected to a cortical neural network with modular functional architecture and connectivity binding functionally distinct neuronal subpopulations from other cortical areas into motor circuit modules at several hierarchical levels [1,2,3,4]. These functional modules display a hierarchy of interleaved circuits connecting, via inter-neurons in the spinal cord, to visual and auditory sensory areas, and to motor cortex, with feed-back loops and bilateral communication with the supraspinal centers [4,5]. This enables a 'from local to global' functional organization with plastic connectivity patterns that are correlated with specific behavioral variations such as variations in motor output or grip force. This functional connectivity fulfills an important adaptive role and ensures that humans are able to reliably grasp and manipulate objects in the physical world under constantly changing conditions in their immediate sensory environment. In an exploratory study, a wireless sensor system was exploited to measure human grip force under varying visual and auditory sensory input conditions to probe the extent to which the grip force variations may reveal adaptive changes in somatosensory-visual-auditory interactions.

## 2. Materials and Methods

A specific wearable sensor system in terms of one glove for each hand with inbuilt Force Sensitive Resistors (FSR), developed in our lab, was used. Hardware and software configurations are described in detail elsewhere here [6,7]. For the details, see: https://www.mdpi.com/1424-8220/19/20/4575/htm.
The glove was designed to acquire analog voltage signals provided by each FSR every 20 milliseconds (msce) at a 50Hz rate. In every loop of the Arduino running software, input voltages were merged with time stamps and voltage levels in millivolt (mV). This data package was sent to the computer via Bluetooth, and a specific software wrote the data into separate text files for each individual participant. Data from eight men, between 20 and 30 years old, all of them right-handed, were analyzed. Handedness was confirmed individually using the Edinburgh inventory. Participants were healthy volunteers and gave their informed consent in conformity with the Helsinki Declaration. Each participant was tested individually, standing upright, with both eyes open or blindfolded, facing a table on which two handles, weighing one kilogram each, for power grip exercise were placed in alignment with the forearm motor axis. In the blindfolded condition, they were made to probe for the position of the handles with their two hands before grabbing them. All participants were instructed to "grab the handles with your two hands and gently move them up and down for ten seconds as soon as the music starts." A CD player with constant output intensity played soft yoga music in one test session, and hard rock music in another.

## 3. Results

Data from the sensor located on the middle phalanx of the middle finger only are analyzed here, given that this finger provides the major contribution to total hand grip force when lifting objects [8]. With one signal every 20 msec, ten seconds per test condition, two visual and two auditory conditions and eight participants, we have a total of 16 000 data for this sensor, and 500 data per individual participant and test condition.

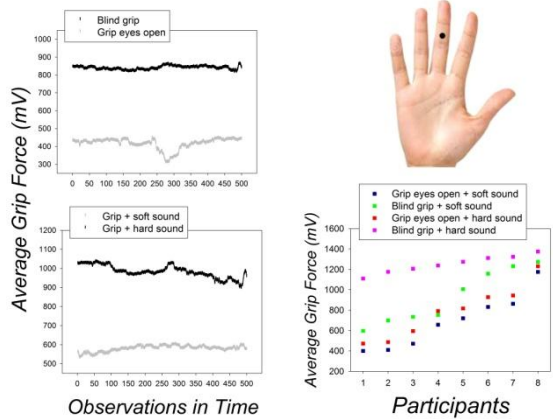

**Fig. 1.** Average middle finger grip force in the two visual conditions (top left), the two auditory conditions (bottom left), and different individuals' average middle finger grip force in each of the four vision-sound factorial combinations (bottom right).

The results (Fig. 1) show clear effects of vision and sound on the middle finger grip forces with blind grip producing stronger forces by comparison with grip eyes open, and hard sound producing stronger grip forces by comparison with soft sounds. The results of 2-Way ANOVA, computed in *Systat* for *Sigmaplot12* on the average middle finger grip forces, are shown in Table 1. The effects of the sound and vision factors are statistically significant, their interaction is not.

**Table 1.** Results of the 2-Way ANOVA with F statistics and their associated probability limits (P).

Two Way Analysis of Variance – Cartesian Design: P8 x V2 x S2
Dependent Variable: Average Grip Forces in Sensor S5

| Factor | DF | SS | MS | F | P |
|---|---|---|---|---|---|
| Vision | 1 | 1006903.602 | 1006903.602 | 18.572 | <.001 |
| Sound | 1 | 339897.445 | 339897.445 | 6.269 | <.018 |
| Vision x Sound | 1 | 104309.710 | 104309.710 | 1.924 | .176 |
| Residual | 28 | 1518065.682 | 54216.631 | | |
| Total | 31 | 2969176.440 | 95779.885 | | |

## 4. Discussion

The middle finger produces most of the grip force necessary for a power grip. Here, it is shown that middle finger grip force adapts spontaneously to changes in the immediate visual and/or sound environment of humans, which translates into a statistically significant, stimulus-driven, change in force output. The spontaneous grip force adaptation is a directly observable consequence, or behavioral correlate, of functional plasticity in the somatosensory brain and connected areas enabling multisensory neural processing where inputs from different sensory modalities are integrated to produce effective sensory-motor activation under changeing conditions. This has implications for robot-driven motor rehabilitation through end-effector systems with adaptive multisensory integration simulating the characteristics of human eye-hand coordination with sensory feedback [9,10,11].

## 5. Conclusions

Human grip force adaptation is a directly observable consequence of somatosensory brain plasticity. Simpler task designs with a well controlled parameter space and well-defined boundary conditions can help optimize automated behaviour through multisensory cueing for motor learning and control through human-system-like adaptive feed-back.

## References

[1]. S. Wilson, C. Moore. S1 somatotopic brain maps. *Scholarpedia*, Vol.10(4), 2015, e-article 8574.
[2]. C. Braun et al. Dynamic organization of the somatosensory cortex induced by motor activity. *Brain*, Vol. 124(11), 2001, pp. 2259-2267.
[3]. S. Arber. Motor circuits in action: specification, connectivity, and function. *Neuron*, Vol. 74(6), 2012, pp. 975-89.
[4]. M. Tripodi, S. Arber. Regulation of motor circuit assembly by spatial and temporal mechanisms. *Curr Opin Neurobiol*, Vol. 22(4), 2012, pp. 615-23.
[5]. T. Weiss et al. Rapid functional plasticity of the somatosensory cortex after finger amputation. *Experimental Brain Research, Vol.* 134(2), 2000, pp. 199-203.
[6]. M. de Mathelin, F. Nageotte, P. Zanne, B. Dresp-Langley. Sensors for expert grip force profiling: towards benchmarking manual control of a robotic device for surgical tool movements, *Sensors*, Vol. 19(20), 2019, e-article 4575.
[7]. B. Dresp-Langley, F. Nageotte, P. Zanne, M. de Mathelin. Correlating Grip Force Signals from Multiple Sensors Highlights Prehensile Control Strategies in a Complex Task-User System. *Bioengineering*, Vol. 7(4), 2020, e-article 143.
[8]. ML. Latash, VM. Zatsiorsky. Multi-finger prehension: control of a redundant mechanical system. *Adv Exp Med Biol*, Vol. 629, 2009, pp. 597-618.
[9]. F. Molteni, G. Gasperini, G. Cannaviello, E. Guanziroli E. Exoskeleton and End-Effector Robots for Upper and Lower Limbs Rehabilitation: Narrative Review. *PM&R*, Vol. 10(9 Suppl 2), 2018, pp. S174-S188.
[10]. LP. Wijesinghe, J. Triesch, BE. Shi. Robot End Effector Tracking Using Predictive Multisensory Integration. *Front Neurorobot*, Vol. 12, 2018, e-article 66.
[11]. T. Schürmann , BJ. Mohler, J. Peters, P. Beckerle. How Cognitive Models of Human Body Experience Might Push Robotics. *Front Neurorobot*, Vol. 13, 2019,e-article1